\documentclass[11pt]{article}
\usepackage[hyperref]{ccl2023-en}
\usepackage{times}
\usepackage{url}
\usepackage{latexsym}
\usepackage{fancyhdr}

\usepackage{multirow}
\usepackage{pifont}
\usepackage{amsmath, amsfonts}
\usepackage{algorithm}
\usepackage{algorithmicx}
\usepackage{algpseudocode}
\usepackage{subfigure}

\usepackage{amsfonts}
\usepackage{amsmath}
\usepackage{graphicx}
\usepackage{multirow}
\usepackage{multicol}
\usepackage{caption}
\usepackage{subfigure}
\usepackage{enumitem}
\usepackage{float}
\usepackage{makecell}
\usepackage{booktabs}
\usepackage{tabu}

\usepackage[T1]{fontenc}
\usepackage[utf8]{inputenc}
\usepackage{longtable}
\usepackage{color}
\usepackage{balance} 
\let\vec\mathbf 
\definecolor{mygray}{gray}{0.6}

\usepackage{tcolorbox}
\usepackage{cleveref}
\usepackage{bbm}
\usepackage{bm}

\newcommand{\dtrain}{\mathcal{D}_{\text{train}}}

\newcommand{\seedset}{\mathcal{S}_{\text{seed}}}

\newcommand{\knn}{$k$-NN}
\newcommand{\ft}{\textsc{FT}}
\newcommand{\pt}{\textsc{PT}}

\newcommand{\jointinf}{\textsc{Union-inf}}
\newcommand{\joint}{\textsc{Union-all}}

\let\vec\mathbf 

\definecolor{right}{RGB}{0,128,96}
\definecolor{wrong}{RGB}{192,0,32}

\title{Revisiting $k$-NN for Fine-tuning Pre-trained Language Models}

\author{
    Lei Li$^{1,2}$, 
	Jing Chen$^{1,2}$, 
	Botzhong Tian$^{1,2}$,
 Ningyu Zhang$^{1,2}$\thanks{$\quad$ Corresponding Author.}\\
	$^1$Zhejiang University \& AZFT Joint Lab for Knowledge Engine, China \\
	\fontsize{11}{10}\selectfont 
	\{leili21, chenjing\_1984, tbozhong, zhangningyu\}@zju.edu.cn \\
}

\date{}

\begin{document}
\maketitle
\begin{abstract}
Pre-trained Language Models (PLMs), as parametric-based \emph{eager learners}, have become the de-facto choice for current paradigms of Natural Language Processing (NLP). In contrast, $k$-Nearest-Neighbor (\knn{}) classifiers, as the \emph{lazy learning} paradigm, tend to mitigate over-fitting and isolated noise. In this paper, we revisit \knn{} classifiers for augmenting the PLMs-based classifiers. From the methodological level, we propose to adopt \knn{} with  textual representations of PLMs in two steps: (1) Utilize \knn{} as prior knowledge to calibrate the training process. (2) Linearly interpolate the probability distribution predicted by \knn{} with that of the PLMs' classifier. At the heart of our approach is the implementation of \knn{}-calibrated training, which treats predicted results as indicators for easy versus hard examples during the training process. From the perspective of the diversity of application scenarios, we conduct extensive experiments on fine-tuning,  prompt-tuning paradigms and zero-shot, few-shot and fully-supervised settings, respectively, across eight diverse end-tasks. We hope our exploration will encourage the community to revisit the power of classical methods for efficient NLP\footnote{Code and datasets are available in \url{https://github.com/zjunlp/Revisit-KNN}.}.
\end{abstract}

\section{Introduction}
\label{intro}

%
%
\cclfootnote{
    %
    %
    \hspace{-0.65cm}  

}

\begin{figure}[!htb] 
\centering 
\includegraphics[scale=1.2]{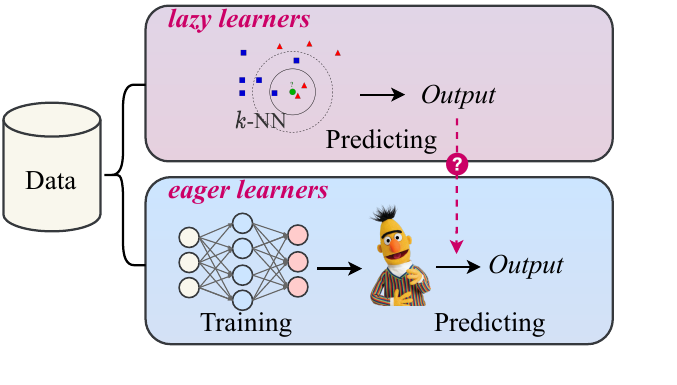} 
\caption{Revisiting how does a lazy learner ($k$-NN) help the eager learner (PLM).
} 
\label{fig:motivation}
\vspace{-0.5cm}
\end{figure}

Pre-trained Language Models (PLMs)~\cite{radfordimproving,Devl2019bert,raffel2020t5}
have shown superior performance across a wide range of language-related downstream tasks~\cite{kowsari2019text,nan2020reasoning}.
Afterward, the conventional paradigm \emph{fine-tuning}, which extends extra task-specific classifiers on the top of PLMs, has been proposed to apply PLMs for downstream tasks.
Recently, a new paradigm called prompt-tuning, which originated from GPT-3~\cite{brown2020language}, has been introduced and has shown better results for PLMs on few-shot and zero-shot tasks. Fine-tuning has proved to be effective on supervised tasks and is widely used as the standard method for natural language processing (NLP).
Despite the effectiveness of adapting PLMs, parametric-based \emph{eager learners}~\cite{friedman2017elements}, like PLMs with neural networks, require estimating the model parameters with an intensive learning stage. Besides, Training a large PLM model can require significant computing resources and energy, which have negative environmental consequences. As a result, there has been a growing interest in developing more efficient and sustainable methods for training and deploying PLMs.




A stark contrast to PLMs is the \knn{} classifier: 
a simplest machine learning algorithm that does not have a training phase but simply predicts labels based on the nearest training examples instead. 
NLP researchers~\cite{DBLP:conf/iclr/KhandelwalLJZL20,he2021efficient} have found that \knn{} enable excellent unconditional language modeling~\cite{DBLP:conf/iclr/KhandelwalLJZL20,he2021efficient} during test phrase.
According the definition in \cite{friedman2017elements}, \knn{} is actually a \emph{lazy learner} that can avoid over-fitting of parameters~\cite{boiman2008nbnn} and effectively smooths out the impact of isolated noisy training data~\cite{orhan2018simple}.
Though \knn{} has the above advantages, previous works only leverage \knn{} for testing, and there is no systematic examination of the full utilization of \knn{} for PLMs.
To this end, we have conducted a comprehensive and in-depth empirical study of the \knn{} classifier for natural language understanding (NLU). Our approach involves leveraging the predictive results of a \knn{}  classifier and augmenting conventional parametric PLM classifiers in two steps:
(1) We explore the role of \knn{}  as prior knowledge for calibrating training by using \knn{}  results as an indicator of easy vs. hard examples in the training set;
(2) During inference, we linearly interpolate probability distributions with the PLM's predicted distributions to make the final prediction;
(3) We conduct extensive experiments with fine-tuning  in fully-supervised, few-shot and zero-shot settings, aiming to reveal the different scenarios where \knn{} is applicable.
We hope this work can open up new avenues for improving NLU of PLMs via \knn{} and inspire future research to reconsider the role of ''old-school`` methods.

\section{Related Work}

\paragraph{\knn{} in the era of PLMs.}
The $k$-Nearest Neighbor (kNN) classifier is a classic non-parametric algorithm that predicts based on representation similarities. While kNN has lost some visibility compared to current deep learning approaches in recent years, it has not fallen off the radar completely. In fact, kNN has been used to enhance pre-trained language models (PLMs) in various tasks, such as unconditional language modeling \cite{DBLP:conf/iclr/KhandelwalLJZL20,he2021efficient}, machine translation \cite{khandelwal2020knnmt,DBLP:conf/aaai/GuWCL18}, and question answering \cite{DBLP:conf/emnlp/KassnerS20}.
Most recently, ~\cite{alon2022neurosymbolic,meng2021gnnlm} further respectively propose automaton-augmented and GNN-augmented retrieval to alleviate the computationally costly datastore search for language modeling. However, previous researchers~\cite{he2021efficient,khandelwal2020knnmt,DBLP:conf/emnlp/KassnerS20,knnbert,meng2021gnnlm,alon2022neurosymbolic,DBLP:journals/corr/abs-2201-05575} mainly focus on generative tasks or adopt simple interpolation strategies to combine \knn{} PLMs only at test time. \cite{knn_prompt} propose to leverage \knn{} for zero-shot inference. 
\paragraph{Revisiting \knn{} for PLMs.} Unlike them, we focus on empirically demonstrating that incorporating \knn{} improves PLMs across a wide range of NLP tasks in fine-tuning and prompt-tuning paradigms on various settings, including the fully-supervised, few-shot and zero-shot settings.
Note that our work is the first to comprehensively explore \knn{} during both the training and inference process further for fruitful pairings: in addition to the approaches mentioned above, we propose to regard the distribution predicted by \knn{} as the prior knowledge for calibrating training, so that the PLM will attend more to the examples misclassified by \knn{}.







\section{Methodology}\label{sec:method}
\begin{figure*}[!t]
\centering
\includegraphics[scale=1.0]{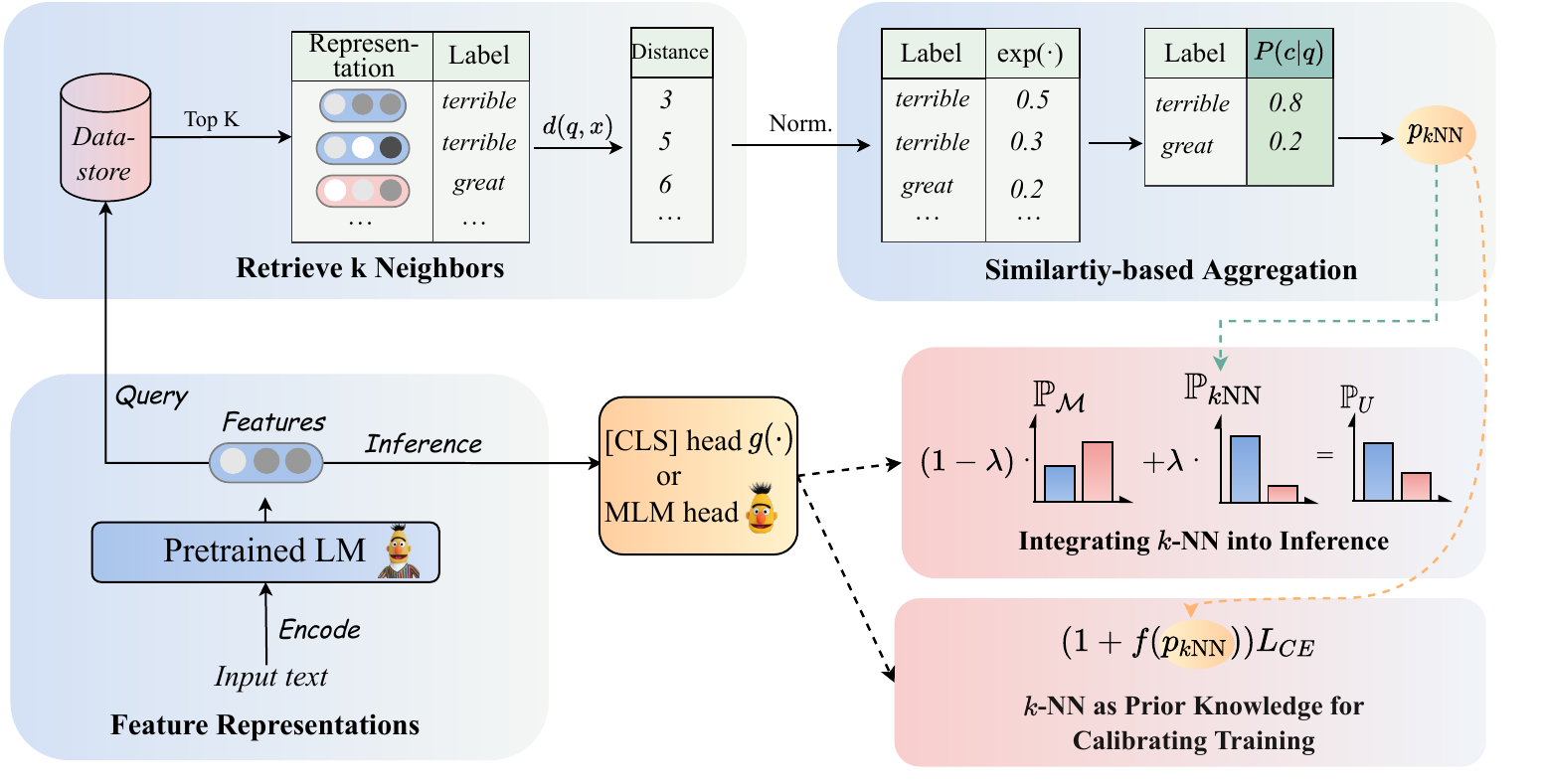}
\caption{Overview of incorporating \knn{} for PLMs }
\label{fig:arc}
\end{figure*}

The overall framework is presented in  \Cref{fig:arc}.
We regard the PLM as the feature extractor that transforms the input textual sequence $x$ into an instance representation $\vec{x}$ with dimensions $D$.
We revisit \knn{} in \S \ref{subsec:knn} and then introduce our method to integrate \knn{} with tuning paradigms in \S \ref{subsec:joint}.

\subsection{Nearest Neighbors Revisited}
\label{subsec:knn}
Given the  training set of $n$ labeled sentences  $\{{x}_1, \dots, {x}_n\}$ and a set of target labels $\{y_1, \dots, y_n\}$, $y\in [1, C]$, the \knn{} classifier can be illustrated in the next three parts:

\paragraph{\textbf{Feature Representations}} 
For \knn{}, we firstly have to collect the corresponding set of features $\mathcal{D} = \{\vec{x}_1, \dots, \vec{x}_n\}$ from the training set.
Concretely, we assign $\vec{x}$ with the embedding of the {\tt[CLS]}  token of the last layer of the PLM for the fine-tuning procedure. More specifically, we define the feature representations as follows:
\begin{equation}
\begin{split}
\vec{x}= 
\mathbf{h}_\texttt{[CLS]}, 
\end{split}
\label{eq:fp}
\vspace{-0.5cm}
\end{equation}
The feature representation $\vec{q}$ of a query example $x_q$ also follows the above equation.


\paragraph{\textbf{Retrieve $k$ Neighbors}}
Following the commonly practiced in \knn{} ~\cite{friedman2017elements,wang2019simpleshot},
we pre-process both $\vec{q}$ and features in the training set $\mathcal{D}$ with $l2$-normalization.
We then compute the similarity between the query $\vec{q}$ and each example in $\mathcal{D}$ with
Euclidean distance as : $d(\vec{q}, \vec{x})$, $\forall \vec{x} \in \mathcal{D}$, where $d(\cdot, \cdot)$ is the Euclidean distance calculation function.
According to the similarity, we select the top-$k$ representations from $\mathcal{D}$, which are the closest in the distance to $\vec{q}$ in the embedding space.

\paragraph{\textbf{Similarity-based Aggregation}}
Let $\mathcal{N}$ donate the set of retrieved top-$k$ neighbors, and 
$\mathcal{N}_y$ be the subset of $\mathcal{N}$ where the whole examples have the same class $y$. 
Then the \knn{} algorithm converts the 
top-$k$ neighbors to $\vec{q}$ and the corresponding targets into a distribution over $\mathcal{C}$ labels. 
The probability distribution of $\vec{q}$  being predicted as $c$ is:
\begin{equation}
    \label{eq:knn_prob}
    p_{k\text{NN}}(c \vert \vec{q}) = \frac{\sum_{\vec{x}\in \mathcal{N}_y}  \exp{} \left(-d(\vec{q}, \vec{x}) / \tau \right)}{\sum_{y \in C} \sum_{\vec{x}\in \mathcal{N}_y}  \exp{} \left(-d(\vec{q}, \vec{x}) / \tau \right)},
\end{equation}
where $\tau$ is the hyper-parameter of temperature.


\subsection{Comprehensive Exploiting  of \knn{}}
\label{subsec:joint}
In this section, we propose to comprehensively leverage the \knn{}, the representative of \emph{lazy learning}, to augment the PLM-based classifier.

\paragraph{\textbf{Role of {\knn{}} as Prior Knowledge for Calibrating Training.}}
As \knn{} can easily make predictions for each query instance encountered without any training, it is intuitive to regard its predictions as priors to guide the network in focusing on hard examples during the training process of language models.
We distinguish between easy and hard examples based on the results of \knn{}. 
Given the probability distribution $p_{k\text{NN}}$ of $\vec{q}$ being predicted as true label $y$, we propose to
adjust the relative loss for the correctly-classified or misclassified instances identified by \knn{}, in order to reweight the cross-entropy loss $\mathcal{L}_{CE}$.
Specifically, we define the calibrated training loss $\mathcal{L}_J$ as:
\begin{align}
    &\mathcal{L}_U = \left(1 +   f(p_{k\text{NN}})\right)\mathcal{L}_{CE} ,
    \label{eq:joint}
\end{align}
where $f(p_{k\text{NN}})$ donates the modulating factor~\footnote{We specify the $f(p_{k\text{NN}})=(1 - p_{k\text{NN}})^{\gamma}$, and other factors are also alternative.} for calibration. 
We are inspired by Focal-loss~\cite{focal_loss} to employ the modulating factor, while our focus is on exploring the application of \knn{} in the fine-tuning of PLMs.

\paragraph{\textbf{Intergrating \knn{} into Inference}}
Let $\mathbb{P}_{{\mathcal{M}}}$ denote the class distribution predicted by the PLM, and $\mathbb{P}_{k\text{NN}}$ be the class distribution predicted by a \knn{} classifier.
Then, the $\mathbb{P}_{{\mathcal{M}}}$ is reformulates by interpolating the non-parametric $k$ nearest neighbor distribution $P_{k\text{NN}}$ using parameter $\lambda$~\cite{DBLP:conf/iclr/KhandelwalLJZL20}  to calculate the final probability $\mathbb{P}_U$ of the label as:

\begin{equation}
\label{eq:jointinf}
   \mathbb{P}_U = \lambda \mathbb{P}_{k\text{NN}} + ( 1 - \lambda) \mathbb{P}_{{\mathcal{M}}},
\end{equation}
where $\lambda\in [0, 1]$ is an adjustable hyper-parameter.

\section{Experiments}

\begin{table}[!htbp]
    \centering
    \scalebox{1.0}{
    \begin{tabular}{cccc}
\toprule
    Dataset &Type & \# Class & Test Size \\ 
\midrule
      SST-5   & sentiment  & 5 & 2,210 \\
      TREC   & question cls  & 5 & 500 \\
      MNLI & NLI  & 3  & 9,815 \\
      QNLI & NLI  & 2  & 5,463 \\
      BoolQ   &QA   & 2 &3,245  \\
      CB  &  NLI  & 3 & 250 \\
      SemEval   & relation extraction  & 19 & 2,717 \\
      TACREV  & relation extraction  & 42 & 15,509 \\
     \bottomrule
    \end{tabular}
    }
    \caption{Detailed dataset statistics.}
    \label{tab:dataset_stat}
\end{table}

\subsection{Datasets}

We choose a variety of NLP tasks to evaluate our proposed methods, including sentiment analysis task (SST-5~\cite{sst2}), question classification task (TREC~\cite{voorhees2000building_trec}), NLI tasks (MNLI~\cite{mnli} and QNLI~\cite{qnli}),
sentence-pair classification task (BoolQ~\cite{boolq} and CB~\cite{cb} ), 
and information extraction tasks (SemEval~\cite{hendrickx2010semeval} and TACREV~\cite{alt2020tacred}).
We also list a detailed introduction of datasets in Table~\ref{tab:dataset_stat}.

\subsection{Experimental Settings}
\paragraph{\textbf{Compared Baseline Methods.}}
We adopt $\text{RoBERTa}_\text{large}$~\cite{liu2019roberta} as the underline PLM and conduct comprehensive experiments to integrate \knn{} into PLMs. 
We choose the baseline approaches and the variant of our proposed method as follows:
(1) \textbf{\knn{}}: the method described in \S \ref{subsec:knn}, which performs classification directly through nearest neighbor retrieval of instance features without relying on any pre-trained language models (PLMs).
(2) \textbf{FT}: which denotes vanilla fine-tuning with PLMs.
(3) \textbf{FT\_Scratch}: which denotes vanilla PLMs in zero-shot setting.
(4) \textbf{PT}: which denotes prompt-tuning with PLMs, similar to \cite{gao2020making}.
(5) \textbf{\jointinf{}}: a variant of our method, which simply linear interpolate \knn{} and paradigms of PLMs during the test time.
(6) \textbf{\joint{}}: the completeness of our approach, which involves applying \knn{} as prior knowledge for calibrating training and also integrating \knn{} into inference.
\paragraph{\textbf{Settings.}}
We test the above methods in full-supervised, few-shot and zero-shot experiments, we assign different settings, respectively:
(1) \textbf{Full-supervised setting:} We use full trainsets to train the PLMs and as neighbors to retrieve.
(2) \textbf{Few-shot setting:}
We follow LM-BFF \cite{gao2020making} to conduct 16-shot experiment and test the average performance with a fixed set of seeds $\seedset$, across three different sampled $\dtrain$ for each task. In this setting, we use the few-shot training set as \knn{} neighbors to retrieve.
(3) \textbf{Zero-shot setting:} We directly evaluate the vanilla FT and \jointinf{} on the test set \textbf{without training}.
As for \joint{}, we take the prompt tuning~\cite{gao2020making} to tag the pseudo labels on \textbf{unlabeled} trainsets and apply untrained \knn{} in the training and inference.

\begin{table*}[!t]
    \centering
    \scalebox{0.87}{
    \begin{tabular}{l|c|cc|cc|cc|cc|c}
    \toprule
    {\multirow{2}{*}{Shot}} 
    & \multirow{2}{*}{Method}
    & SST-5   & TREC      &  MNLI      &  QNLI      & BoolQ   & CB  & SemEval    & TACREV & AVG \\
     &     &   Acc.     &  F1.              &  Acc.        & Acc.         & Acc.    &  F1.   &  F1.   &  F1.  & Score.\\

    \midrule
       \multirow{4}{*}{Full} 
    & \textbf{\knn{}}             & $35.8$ & $80.0$ & $41.5$ & $57.2$  & $61.4$ & $42.3$ & $2.5$ & $5.3$ & $40.8$ \\
     &  \textbf{\ft{}}            & $59.2$ & $97.8$ & $83.9$ & $89.1$  & $81.7$ & $89.5$ & $89.4$ & $72.5$ & $82.9$ \\
      &  \textbf{\jointinf{}}       
    & $59.5$ & $98.0$ & $84.0$ & $89.2$ & $82.9$ & $89.6$ & $89.2$ & $67.8$ & $82.5$\\
    &  \textbf{\joint{}}            
    & \underline{$60.9$} & \underline{$98.2$} & \underline{$84.2$} & \underline{$90.8$} & \underline{$83.4$} & \underline{$90.5$} & \underline{$89.6$} & \underline{$73.1$} 
    & \underline{$83.8$}  \\    
    \midrule
    \multirow{4}{*}{16} 
    & \textbf{\knn{}}             & $25.6_{2.4}$ & $46.1_{5.0}$ & $33.7_{0.3}$ & $51.6_{1.3}$  & $50.4_{2.6}$ & $40.8_{4.9}$ & $0.5_{0.4}$ & $0.9_{0.3}$  & $31.1$ \\
     &  \textbf{\ft{}}            & $43.3_{0.7}$ & $86.6_{4.7}$ & $44.4_{4.5}$ & $55.3_{3.7}$  & $56.0_{4.2}$ & $68.3_{4.7}$ & $64.1_{2.3}$ & $25.6_{0.3}$ & $55.5$ \\
     &  \textbf{\jointinf{}}       
    & $43.0_{1.2}$ & $86.7_{4.5}$ & $44.5_{4.5}$ & $55.4_{3.4}$ & $55.4_{4.3}$ & $65.6_{4.7}$ & $65.1_{2.1}$ & $30.5_{1.7}$ & $55.8$ \\
     &  \textbf{\joint{}}            
    & \underline{$43.7_{0.5}$} & \underline{$90.0_{3.9}$} & \underline{$51.7_{1.8}$} & \underline{$58.1_{2.7}$} & \underline{$57.6_{2.7}$} & \underline{$69.8_{4.5}$} & \underline{$67.2_{3.3}$} & \underline{$32.1_{3.1}$} 
    & \underline{$58.9$}\\

   \midrule
       \multirow{4}{*}{0} 
    &{\textbf{\ft{}\_Scratch}}            
    & $23.8$ & $22.6$ & $31.6$ & $49.5$  & $37.8$ & $21.5$ & $8.2$ & $0.1$
    & $24.4$ \\
    &  \textbf{\pt{}}            & $36.7$ & $38.2$ & $50.9$ & $50.8$  & $62.2$ & $39.7$ & $10.9$ & $1.1$ & $36.3$ \\
   &  \textbf{\jointinf{}}       
    & \underline{$51.6$} & \underline{$82.4$} & \underline{$67.5$} & \underline{$67.4$} & \underline{$62.9$} & \underline{$56.9$} & \underline{$11.8$} & \underline{$3.2$}
     & \underline{$50.5$}\\
    &  \textbf{\joint{}}            
    & $35.1$ & $38.0$  & $53.7$ & $50.4$ & $62.4$ & $50.3$ & $11.3$ & $1.4$ & $37.8$ \\

    \bottomrule
    \end{tabular}
    }
   \caption{Results on eight NLP tasks across the fully-supervised, few-shot (16-shot) and zero-shot settings. For the 16-shot setting, we provide the mean and standard deviation across three different random seeds. Scores that are marked with an \underline{underline} signify the best results among all methods.}
   \label{tab:exp_main}
    \vspace{-1em}
\end{table*}

\subsection{Hyper-parameter Settings}
\label{app:exp_settings}
We report the hyper-parameters   in Table~\ref{tab:app_exp_settings}. For the GLUE and SuperGLUE datasets, we follow LM-BFF\footnote{\url{https://github.com/princeton-nlp/LM-BFF}} to construct templates and verbalizer for prompt-tuning. While for RE datastes, we follow KnowPrompt~\cite{chen21knowprompt} to construct templates and verbalizer. We utilize Pytorch to conduct experiments with 1 Nvidia 3090 GPUs. 
We used the AdamW optimizer for all optimizations, with a linear warmup of the learning rate
followed by a linear decay over the remainder of the training. The hyper-parameter settings used in our experiments are listed below.

\begin{table}[!htbp]
    \centering
    \scalebox{1.0}{
    \begin{tabular}{c|c}
\toprule
      Hyper-parameter   & Value \\
\midrule
   maximum sequence length      &  \{128, 256\} \\
   max training step & 1000 \\
   evaluation step & 100 \\
   learning rate & \{1e-5, 2e-5, 5e-5\} \\
   batch size & 8 \\
   gradient accumulation step & \{2, 4, 8\} \\
   adam epsilon & 1e-8 \\
   $k$ & \{16, 32, 128\} \\
   $\lambda$ & \{0.1 : .1 : 0.9\} \\
   $\tau$ & \{0.01, 0.1, 1, 10\} \\
\bottomrule
    \end{tabular}}
    \caption{Hyper-parameter settings.}
    \label{tab:app_exp_settings}
\end{table}

\subsection{Main Results}

\paragraph{\textbf{\knn{} features result in performance gains.}}
We compare the specific results with baseline models and provide comprehensive insights of \knn{} on different paradigms and different settings. 
The results as shown in Table 1. Leverage \knn{} features results in performance gains in both few-shot and fully-supervised settings. In the zero-shot setting, PT-based methods outperform FT-based and \knn{} features further enhance the performance of PT-based methods, which demonstrates that it is flexible and general to integrate \knn{} for PLMs. 


\begin{figure*}[!t] 
\centering 
\includegraphics[width=1\textwidth]{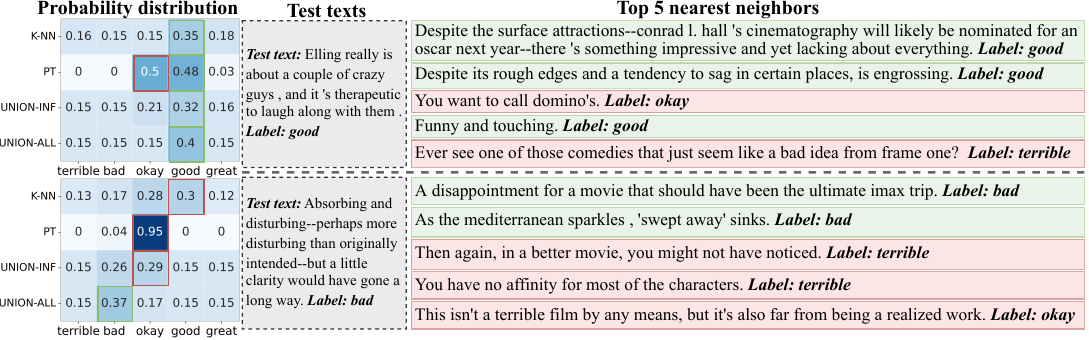} 
\caption{Case analysis to show how \knn{} benefits the prediction of PLMs. We illustrate the test texts, the predicted probability distribution, and the top-5 nearest neighbors from the 16-shot training set of the SST-5 dataset.} 
\label{fig:case}
\end{figure*}

\paragraph{\textbf{Calibrating training vs. Incorporating into inference.}}\quad 
It is necessary to study the different application scenarios of incorporating \knn{} during the training and testing phases. 
From Table~\ref{tab:exp_main}, we observe the following:
(1) Leveraging \knn{} during the test phrase is especially helpful for  the zero-shot setting.
While \textbf{\joint{}} performs worse due to the noise brought from the pseudo-labels on unsupervised data.
(2) \textbf{\jointinf{}} is not doing as well in the fully-supervised and few-shot setting.
In contrast, \textbf{\joint{}} outperforms \textbf{\jointinf{}} in these settings, especially in the few-shot setting.
These findings reveal to us the applicable scenarios of incorporating \knn{} and inspire further studies to utilize \knn{} classifier more practically for efficient NLP.



\subsection{Analysis}
\paragraph{Q1: How does the lazy learner benefit eager learner?}
To further understand how does the \emph{lazy learner} (\knn{}) benefit the  \emph{eager learner} (PLM), we manually check cases in which \knn{}, PT, \jointinf{} and \joint{} produce different results.
As shown in the example of the upper row of Figure~\ref{fig:case}, 
\knn{} and \joint{} predict correctly 
when PT fails. 
This result is because \joint{} produces a more confident probability for the correct class via calibrating the attention on the easy vs.~hard examples identified by the \knn{} classifier.
Note that the bottom row shows that \joint{} predicts correctly even when \knn{} predicts wrongly, possibly due to the robustness of \knn{} calibration.

\paragraph{Q2: Does the similarity metric matter?}
In the above experiments, we mainly utilize negative ${L2}$ distance to measure the similarity between the query $\vec{q}$ and the instance representation of the data store. 
It is intuitive to estimate the impact of different similarity metrics, such as cosine similarity. 
Thus, we present the performance of \joint{} using both metrics with the same hyperparameters as below.

\begin{center}
\begin{tabular}{c|c|c}
Similarity Metric & ${L2}$ & ${cos}$ \\
\midrule
16-shot SST-5 (\%) &\textbf{43.7}  & 42.8\\
16-shot TREC (\%) &\textbf{90.0}  & 89.4\\
16-shot QNLI (\%) &\textbf{58.1}  & 57.2\\
\end{tabular}
\end{center}

We can find that \joint{} with cosine distance achieves nearly the same performance as those trained with ${L2}$, revealing that our \joint{} is robust to the similarity metric. 

\begin{figure}[!htbp]
\centering
\includegraphics[scale=1.0]{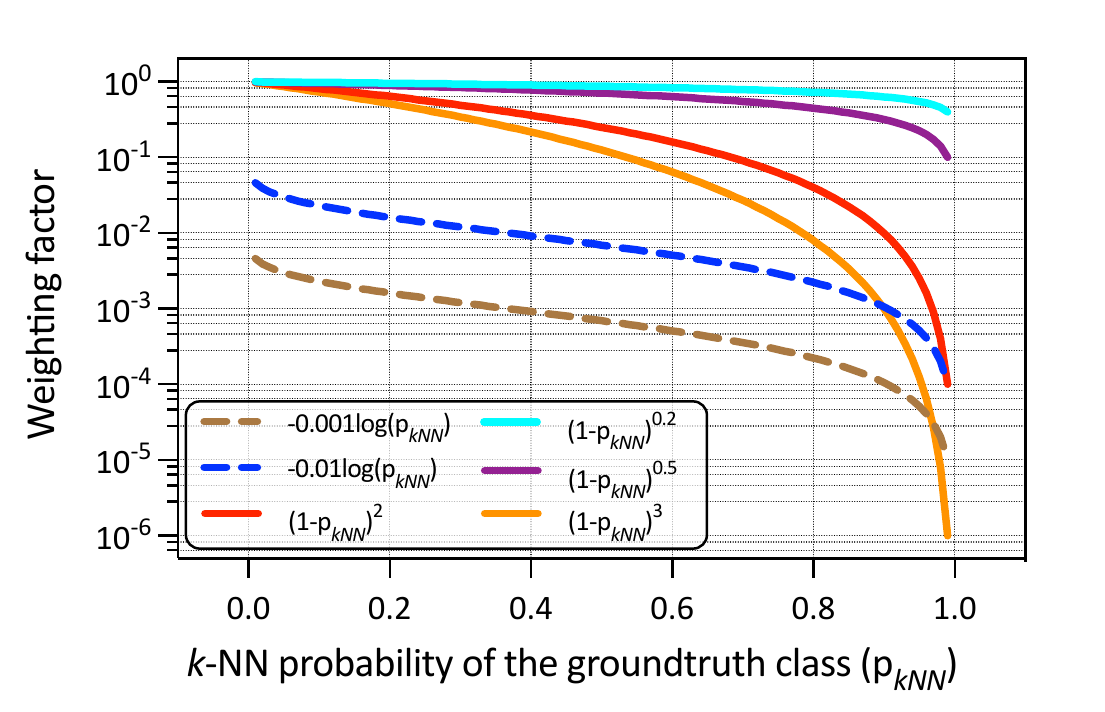}
\caption{
Comparison between the modulating factors NLL and Focal.
}
\label{fig:ana_weight}
\end{figure}

\paragraph{Q3: How dose the modulating factor $f(p_{k\text{NN}})$ works?}
Since we adopt focal loss (Focal) as the modulating factor for main experiments, we further explore other functions as modulating factors, such as negative log-likelihood (NLL).
As shown in Figure~\ref{fig:ana_weight}, we visualize two modulating factors with different settings of $\alpha$ and $\gamma$, where $\alpha$ donates a scalar that represent the proportion of the term of NLL, and $\gamma$ is the exponential coefficient for Focal.
We can find that NLL and Focal produce large weights for the misclassified examples, demonstrating the diversity of modulating factor selection.

\section{Limitations}
We only explore leveraging the training data for \knn{} search, while various external domain data are also suitable for k-nearest neighbor retrieval.
Moreover, incorporating \knn{} also faces the following limitations: (1) the requirement of a large memory for retrieval; (2) hyper-parameters (such as $\lambda$ and $\alpha$) used for retrieval have an impact on the performance of model training; (3) if the number of nearest neighbors $k$ is too large, it will also affect the efficiency.

\section{Conclusion and Future Work}
In this paper, we propose a novel method to enhance PLM-based classifiers using \knn{}. Specifically, we introduce a calibration process and linear interpolation of inference phrases to effectively integrate \knn{} into the training pipeline. To evaluate the effectiveness of our approach, we conduct a comprehensive and in-depth analysis of the role of \knn{} in various NLU tasks and tuning paradigms. Our results demonstrate that the integration of \knn{} is flexible and can significantly enhance the performance of large models. Future work should explore the combination of \knn{} and LLMs such as (1) Inject external knowledge into the LLMs with \knn{}. Specifically, \knn{} can be used to retrieve relevant knowledge from an external database during the reasoning process, which can help correct errors and reduce the prevalence of gibberish output and factual errors that are common in LLMs.
(2) Retrieve contextual information to enhance LLMs. \knn{} algorithms can automatically retrieve relevant information based on the input sentence, such as instructions or other relevant context.
(3) Augment the training data for LLMs. \knn{} is a powerful tool for identifying similar instances in a large dataset, which can help overcome the limitations of data scarcity and improve the performance LLMs.

\clearpage



\bibliographystyle{ccl}
\balance
\bibliography{sample-base}

\end{document}